\documentclass[twocolumn,letterpaper,10pt]{article}
\usepackage{iasted}
\usepackage{times}
\usepackage{graphicx}
\usepackage{subfig} 
\usepackage{amsmath}
\usepackage{algorithm}
\usepackage{algorithmic}

\begin{document}
\date{}

\vspace{-0.2cm}
\title{QCQP-Tunneling: Ellipsoidal Constrained Agent Navigation}

\author{Sanjeev Sharma\\
Electrical Engineering Department,\\
Indian Institute of Technology Roorkee,\\
Roorkee, 247667, India\\
email: sanshuee@iitr.ernet.in
\thanks{Author earned his B.Tech in Electrical Engineering from Indian Institute of Technology Roorkee, 2011, and did this research as an undergraduate student.\newline\newline This paper was accepted in the proceedings of Second IASTED International Conference on Robotics, 2011; DOI: 10.2316/P.2011.752-010}}
\maketitle
\thispagestyle{empty}

\noindent
{\bf\normalsize ABSTRACT}\newline
{This paper presents a convex-QCQP based novel path planning algorithm named ellipsoidal constrained agent navigation (ECAN), for a challenging problem of online path planning in completely unknown and unseen continuous environments. ECAN plans path for the agent by making a tunnel of overlapping ellipsoids, in an online fashion, through the environment. Convex constraints in the ellipsoid-formation step circumvent collision with the obstacles. The problem of online-tunneling is solved as a convex-QCQP. This paper assumes no constraints on shape of the agent and the obstacles. However, to make the approach clearer, this paper first introduces the framework for a point-mass agent with point-size obstacles. After explaining the underlying principle in drawing an ellipsoid tunnel, the framework is extended to the agent and obstacles having finite area (2d space) and finite-volume (3d-space).} \vspace{2ex}
   
\noindent
{\bf\normalsize KEY WORDS}\newline
{Path Planning, Mobile Robot/Vehicle Navigation, Convex Optimization, Continuous Environment.}
\vspace{-0.2cm}
\section{Introduction}
Path planning is a vital component of autonomous navigation. Despite the recent success of DARPA Urban Challenge, path planning in completely unknown, unseen and continuous spaces is still a challenging problem. This paper presents a convex-QCQP ([1]) formulation based novel algorithm (ECAN) for path planning in completely unknown, unseen and continuous 2D and 3D spaces. ECAN generates a path by creating a tunnel of overlapping ellipsoids through the environment. Convex constraints in the ellipsoid-formation step ensure obstacle avoidance. ECAN uses a point-cloud representation of surrounding obstacles to plan a path, thereby facilitating direct implementation in real-world as most of the sensors (image, 3D-Lidar etc.) return point cloud representations of surrounding obstacles.

When the environment is known or partially known, RRT \& spline based planners ([2],[3]) and grid-based planners \& re-planners ([4],[5],[6],[7]) can be used. Recent research has focused on extending grid-planners to continuous heading directions ([6],[7],[8]) since discretization often results in suboptimal path even when the environment is known. A reasonable approach to plan in continuous spaces is to use optimization techniques. In recent years, with increasing computational resources, a lot of progress has been made in path planning using the MILP, both in known and partially known environments ([9],[10],[11],[12]). MILP is a powerful mathematical programming tool that can handle collision avoidance constraints and provide an efficient path through the environment. However, computational requirements with MILP based planners grow too high even in known environments with reasonable number of constraints ([11],[12]), making them practically less significant. This paper presents an unknown and unseen spaces path planning algorithm, that scales well with number of obstacles, using convex-optimization ([1]). Convex optimization has been widely accepted as a powerful tool to solve many engineering problems. Convex functions, most importantly, do not have the pitfall of local minima, assuring that the achieved solution always happens to be the optimal. Despite this, direct implementation of convex optimization for path planning, like MILP, is a less explored area. To the best of our knowledge, this is the first ever work using convex optimization (in this case a QCQP) as a main planner for path planning in completely unknown and unseen spaces. ECAN utilizes only the information in the field-of-view (fov) of the agent to plan (online) a path leading to the goal location. Also, unlike most of the literature in path planning, the proposed algorithm (ECAN) does not restrict the agent/vehicle to a point-mass entity and can directly handle the agent having finite-area/volume with any shape. Simulation results show feasibility of ECAN in UGV and UAV (helicopters/fixed-wing planes) path planning. The paper proceeds as follows: sec-\ref{relatedwork} presents related work; sec-\ref{basis-underlying-priniciple} outlines the underlying principle of ECAN with point-agent and point-obstacles; sec-\ref{ECAN-Framework} outlines notations, describes ECAN framework and navigation strategy with point-masses; sec-\ref{finite-agent-architecture} explains finite agent \& obstacle architecture; and sec-6 and 7 present experiments and conclusion \& future work respectively. 
%
%
\section{Related Work}
\label{relatedwork}
A 2D Tunnel-MILP algorithm is presented in [12] for planning a path through a sequence of pre-decided convex polytopes. However, this method is computationally intensive, and it works only in known \& seen environments. A similar work by Blackmore [13] proposes UAV path planning with stochastic uncertainty (wind) using a two-stage optimization algorithm. At the first stage, an MILP generates a feed forward map (reference control) which is then used to design a feedback (path planning with uncertainty) control, at second stage, for a linear dynamical UAV model using \textit{chance constrained} formulation (solving SOCP, convex program). The algorithm models uncertain risks, which the present paper doesn't model, but to solve the first-stage optimization problem a deterministic and completely known environment model is necessary. Moreover, as far as UAV navigation is concerned, the paper used a constant altitude model. Vandapel \textit{{et}.{al}.} [14] construct a 3D-network of tunnels formed by overlapping spherical bubbles through the cluttered environment. Their approach is again limited to known and seen environment and is basically a graph based approach using A* ([15]) search to decide an optimal path to the goal location through the network of tunnels. An SVM (convex program) based path planning approach is presented in [16]. The method uses the environment model, and then creates dummy obstacles around the nominal line (plane, in 3D-space) to guide the SVM to generate a desired path. A control-point insertion based spline fitting algorithm is presented in [17], taking into account the width and height of the robot. Path planning is done for ground navigation in an environment with 3D-objects. The algorithm also considers planning in non-planner surfaces by planning over mesh. An extension of this algorithm to UAV path planning would be a good approach. However, in the present version for ground navigation, the algorithm needs to know the locations of all the obstacles to find control points. In [18], a grid-based path planner plans in completely unknown spaces. The algorithm is based on straight line formalisms, and obstacles are avoided on-the-fly using a table of rules which rotates the line to avoid obstacles. Once a path is planned and the sub-goals are identified, advancing learning and trial-and-error learning improve the already planned path using the (now partially known) environment model and sub-goals. Limitations of this approach are: (i) it suffers form limited heading constraints, unlike ECAN which plans in continuous spaces; and (ii) the method can provide a  resonable path only when the environment becomes (at least) partially known, while ECAN directly provides a reasonable (if not optimal, since the space is unknown and useen) path to the goal location. [19] presents an iterative MILP formulation for UAV (again constant altitude) path planning in partially known environment (with discovered obstacles known to be static) and performs tests on real flights to complete a coordinated mission. However the algorithm suffers from the computational requirements of MILP ([11],[12]). Also, to ensure safe navigation, UAV was provided with a backup trajectory and the online planner avoids obstacles following this trajectory. [20] presents a 3D-MILP planner, but requires a pre-specified path to the goal location, and is therefore limited to seen spaces. 
%
%
\section{ECAN: Underlying Planning Principle}
\label{basis-underlying-priniciple}
\begin{figure}[t]
\begin{center}
\subfloat[]{\label{fig:working-principle1}\includegraphics[trim =45mm 90mm 46mm 151.8mm, clip, width=2in]{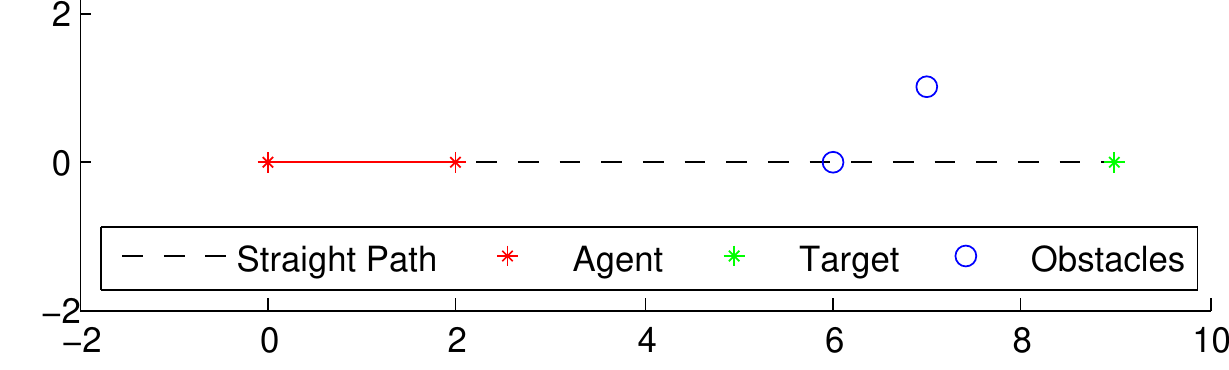}}\\
\subfloat[]{\label{fig:working-principle2}\includegraphics[trim =65mm 105mm 56mm 152mm, clip, width=1.5in]{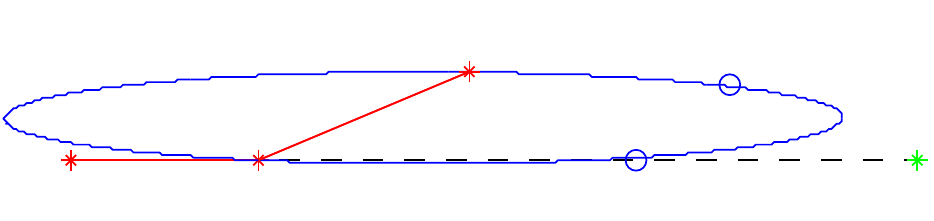}}
\subfloat[]{\label{fig:working-principle3}\includegraphics[trim =65mm 105mm 56mm 152mm, clip, width=1.5in]{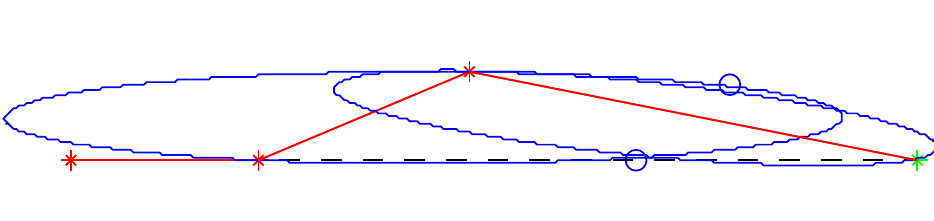}}
\end{center}
\caption{Basic working principle of ECAN.}
\label{fig:working-principle}
\end{figure}

Before beginning to describe the convex formulation in ECAN algorithm, a basic underlying principle is explained. This makes the understanding of later sections easier. For simplicity, this section considers 2D-navigation, a point-agent and static point-obstacle environment. The principle is also valid for 3D-planning. Assume an agent at $(0,0)$ trying to reach at $(9,0)$, see fig-\ref{fig:working-principle1}. After navigating to $(2,0)$ agent discovers an obstacle at $(6,0)$. At this step an ellipsoid is formed, taking into account, the agent's location, the obstacle's location and the target location (fig-\ref{fig:working-principle2}). Since the goal location is not lying on the boundary of the ellipsoid (indicating the presence of an obstacle), agent finds a navigation direction inside the ellipsoid (to avoid obstacle) and navigates to the safe location on the boundary of the ellipsoid (fig-\ref{fig:working-principle2}). After navigating to the boundary, agent discovers another obstacle at $(7,1)$. Again an ellipsoid is formed, and since the target location is on the boundary, agent navigates to this location, reaching the destination (fig-\ref{fig:working-principle3}). Later sections describe the process of ellipsoid formation (sec-\ref{QCQP-Formulation}), finding the navigation direction ($z_n$) inside the ellipsoid and the step-length ($l_n$) to move inside the ellipsoid (sec-\ref{inside-point-2d} and \ref{finite-agent-architecture}). Also the environment may be dynamic and moreover, in case of finite-area/volume obstacles, the agent may discover only a part of the obstacle in its field-of-view instead of the full obstacle. Therefore, due to the associated uncertainty, navigating to a safe location on the boundary of the ellipsoid may not be possible. This is tackled by finding a safe step-length to move inside the ellipsoid (sec-\ref{Quadratic-Program}).
%
%
%
%
\section{Ellipsoidal Constrained Agent Navigation}
\label{ECAN-Framework}
 
This section presents ECAN algorithm and the underlying convex programming framework for path-planning of a point-agent navigating in an environment polulated with randomly scattered point-obstacles. The work utilizes the SDP framework for separating two data-sets by a quadratic surface [1], in this case an ellipsoid. This paper assumes that the agent has a finite field-of-view (fov) and it cannot see beyod its fov. Agent's fov is defined by parameters $(r,\theta)$ in 2D-space and by $(r,\theta,\phi)$ in 3D space (using polar coordinate system), where $r\in(0,R_\textit{fov}]$, $\theta\in[-\theta_\textit{fov},\theta_\textit{fov}]$ and $\phi\in[-\phi_\textit{fov},\phi_\textit{fov}]$. Parameters ($R_\textit{fov},\theta_\textit{fov},\phi_\textit{fov}$) restrict the agent's fov. Obstacles discovered at time step $t$ are the obstacles lying in the agent's fov. Also, agent has a local coordinate system represented as $\zeta^2_a$ in 2D navigation and $\zeta^3_a$ in 3D-navigation. Further details regarding the fov and implementation are in sec-\ref{obstacle-sampling}, this section discusses the working principle of ECAN algorithm. Throughout the paper, $z_{a}^t\in{R^2}$ ($R^3$, for 3D-space) represents the agent's location at time $t$, $z_{o_i}^t\in{R^2}$ ($R^3$, for 3D-space), $i\in\{1,...,k\}$ represents location of the $i^{th}$ point-obstacle at time step $t$ and $z^t_{g}\in{R^2}$ ($R^3$, for 3D-space) denotes the goal location that the agent is aiming for at time $t$. An ellipsoid at time $t$ is represented as $\Psi^t(P,q,r)=\{x|x^TP^tx+x^Tq^t+r^t\leq{0}\}$, with its interior $\textbf{int}(\Psi^t)=\{x|\Psi^t(P,q,r)<0\}$ and $\Psi^t(z)=z^TP^tz+z^Tq^t+r^t$ represents value of point $z$ w.r.t. $\Psi^t$. A set of all $n\times{n}$ positive definite symmetric matrices is denoted as $S^n_{++}$. For a 2D-planning problem, $x\in{R^2}; P\in{S^2_{++}}, q\in{R^2}, r\in{R}$ and for 3D, $x\in{R^3}, P\in{S^3_{++}}, q\in{R^3}, r\in{R}$, $v^T$ represents transpose of vector $v$, $|P|$ denotes determinant of a matrix $P{\in}S^n_{++}$, and $|q|$ denotes absolute value of $q\in{R}$. The minimum and maximum Eigenvalues of $P$ are denoted as $\lambda_{min}(P)$ and $\lambda_{max}(P)$ respectively. Sec-\ref{QCQP-Formulation} explains QCQP formulation using the point agent and the point obstacles; sec-\ref{inside-point-2d} explains navigation inside the ellipsoid for a point-agent; and sec-\ref{finite-agent-architecture} extends 
the approach to the finite-area/volume agent navigating in spaces populated with the finite-area/volume obstacles.
%
%
\begin{figure}[t]
\begin{center}
\subfloat[]{\label{fig:importance_f2}\includegraphics[trim =90mm 128mm 50mm 100mm, clip, width=1.8in]{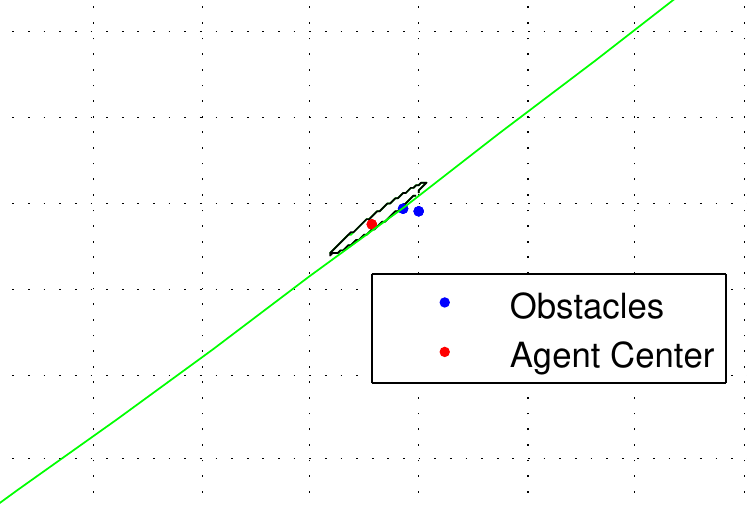}}
\subfloat[]{\label{fig:table_f2}\includegraphics[trim = 30mm 200mm 80mm 25mm, clip, width=2in]{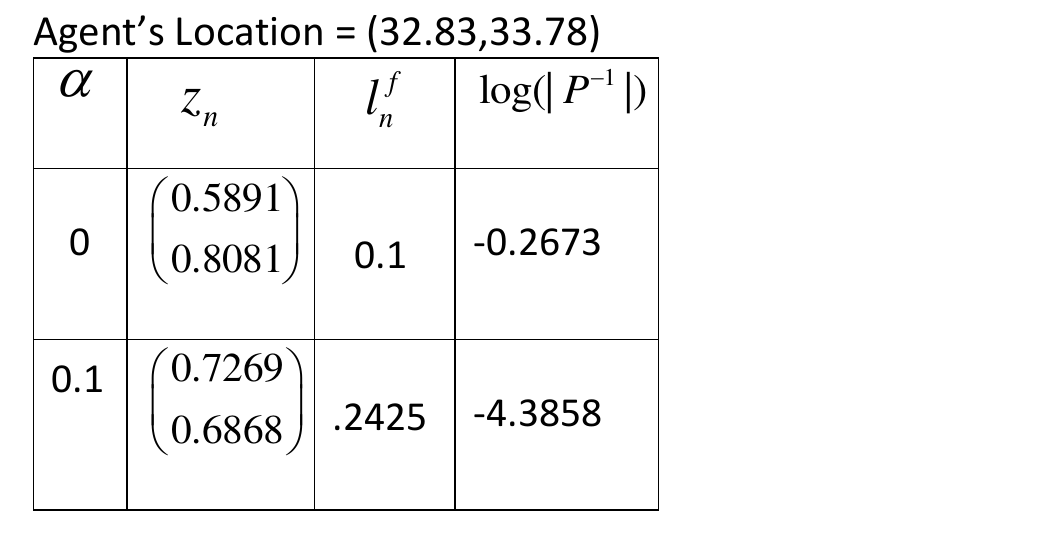}}
\end{center}
\caption{(a) resulting ellipsoids with $f_2$ (black), and w/o $f_2$ (green) in convex problem-(\ref{ccf1f2f3}); (b) different parameters due to resulting ellipsoid. Ellipsoid without convex objective $f_2$ is too large to be depicted completely.}
\end{figure}
%
%
\subsection{ECAN Path Planning: Convex-QCQP Tunneling} 
\label{QCQP-Formulation}

ECAN is a 3-step online algorithm for planning in continuous spaces. The 3-steps are: (i) determining a feasible ellipsoid at time $t$; (ii) finding a navigation direction for navigating inside the ellipsoid; and (iii) determining the step-length to move in that direction. Each of the 3-steps requires solving a convex optimization problem. This section explains the first-step of ECAN, i.e. finding a feasible ellipsoid. An ellipsoid forming at time $t$ ($\Psi^t$), parameterized by $(P^t,q^t,r^t)$, should satisfy 3 constraints: (i) agent lies inside the ellipsoid ($\Psi(z_a^t)<0$); (ii) obstacles lie outside the ellipsoid ($\Psi(z^t_{o_i})>0, i=\{1,...,k\}$); and (iii) goal location can either be outside or on the boundary of ellipsoid ($\Psi(z^t_{g})\geq{0}$). These three constraints ensure a navigable space (if it exists) for the agent even in a highly cluttered environment and a constraint $P\in{S^n_{++}}$ results in an ellipsoid surface. Next, convex objective functions are designed such that the resulting Convex-QCQP motivates the agent to move towards the goal location.

To lure the agent to move towards goal location, the ellipsoid boundary should be as close as possible to the goal location. This is equivalent to the principal axis of resulting ellipsoid being aligned with the goal location (in ideal case with no obstacle, goal location lying in a ray from center of the ellipsoid in the direction of principal axis). This is done by adding a convex penalty term ($f_1$) in the objective which is $zero$ iff the goal location lies on the boundary of the resulting ellipsoid. A penalty function $f_2$ is added which minimizes distance of current location of the agent from the boundary of resulting ellipsoid. At first this may seem counter-intuitive, but eventually it leads to a stable ellipsoid. Without $f_2$, $\Psi^t$ can grow unboundedly large in any direction to satisfy the convex constraints. This can result in a very small navigable distance in the feasible navigation direction inside the ellipsoid. Fig-\ref{fig:importance_f2} shows two ellipsoids, one determined using $f_2$ (black) and other (green) without $f_2$ (see, table in fig-\ref{fig:table_f2} and sec-\ref{inside-point-2d} for further details). A third penalty function $f_3$ is added to make a locally maximal ellipsoid around the agent, taking into account the surrounding obstacles at time $t$. The resulting Convex-QCQP formulation is (with variables $P^t$, $q^t$, $r^t$),
%
%
\begin{eqnarray}
\label{ccf1f2f3}
\textbf{minimize}& &f_1^{}+{\alpha}f_2^{}+\gamma{f_3^{}}\\
\textbf{subject to}& &\text{ }{{{z^t_a}}^TP^tz^t_a+{q^t}^Tz_a^t+r^t\leq{-1}}\nonumber\\
& &{{z^t_{g}}^TP^tz^t_{g}+{q^t}^Tz^t_{g}+r^t\geq{0}}\nonumber\\
& &{z^t_{o_i}}^TP^tz^t_{o_i}+{q^t}^Tz^t_{o_i}+r^t\geq{1}\nonumber\\
& &i=\{1,...,k\}; P^t\succeq{I}\label{ecan_acsf}\nonumber.
\end{eqnarray}
Here $f_1=||{z^t_{g}}^TP^tz^t_{g}+{q^t}^Tz^t_{g}+r^t||_1$,  $f_2=||{z^t_a}^TP^tz^t_a+{q^t}^Tz^t_a+r^t||^{2}_2$, $f_3^{}=\sum_{i=1}^k\Psi^t(z^t_{o_i})$, $\alpha\in(0,1]$, and $\gamma\in(0,10^{-3}]$. As mentioned earlier, function $f^{}_3$ results in affinity of resulting ellipsoid with the obstacles identified at time $t$; $\gamma$ decides the magnitude of this affinity.
%
%
\subsection{ECAN Path Planning: Navigation Inside Ellipsoid}
\label{inside-point-2d}

After finding a feasible ellipsoid at time step $t$, a direction for navigation inside the ellipsoid needs to be found. The navigation direction $z_n$ should take the agent away from the surrounding obstacles, while should also lead the agent towards the goal location. Function $f_1$ in convex-QCQP tries to align the resulting ellipsoid such that the principal axis points towards the goal location, while $f_3$ distorts this alignment by making the ellipsoid reactive to surrounding obstacles. This section now demonstrates how navigation direction is found using the ellipsoid's Eigenvectors and Eigenvalues. If $\Psi^t(z^t_g)=0$ at time-step $t$, then the navigation direction $z_n$, is simply a (unit) vector from $z^t_a$ to $z^t_g$. However, if $\Psi^t(z^t_g)>0$ then the navigation direction has to be decided. This direction is found by first solving a convex program, which is different for 2D and 3D navigation problems. Following sub-sections describe the method of finding a navigation direction when $\Psi^t(z^t_g)>0$.
\begin{figure}[b]
\begin{center}
\includegraphics[trim = 12mm 95mm 21mm 52mm, clip, width=8cm]{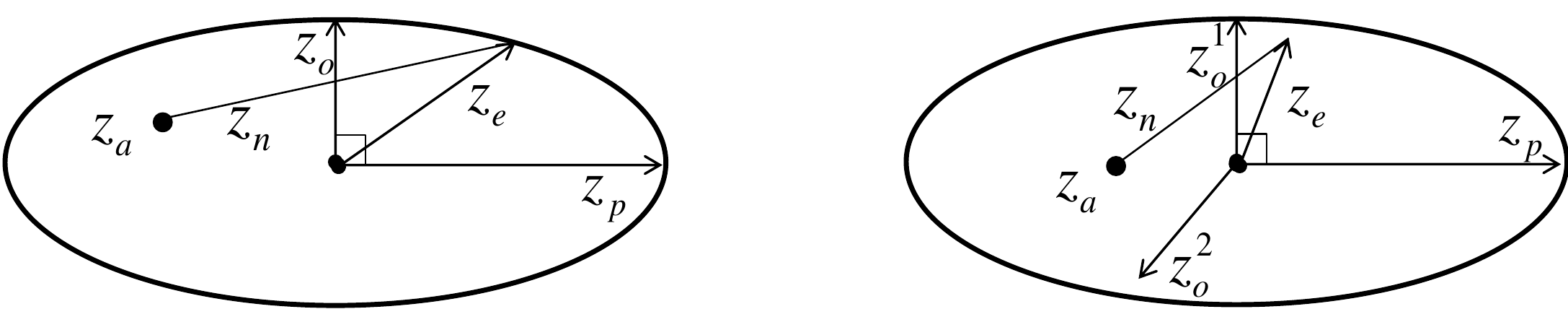}
\end{center}
\caption{Left figure depicts inside ellipsoid vectors $z_p, z_o, z_e$ and $z_n$ for 2D-space and figure on right depicts $z_p, z^1_o,z^2_o,z_e$ and $z_n$ for 3D-space.}
\label{fig:inside_ellip_vectors}
\end{figure}
%
%
%
\subsubsection{2D-ECAN: Avoiding Obstacles}

In 2D case, two direction vectors are defined, the principal direction vector $z_p$ and the obstacle vector $z_o$, which correspond to eigenvectors corresponding to $\lambda_{min}(P)$ and $\lambda_{max}(P)$ respectively. Due to the objective functions $f_1$ and $f_2$ in (\ref{ccf1f2f3}), vector $z_p$ correlates with the $x$-axis of the agent's local coordinate system $\zeta^2_a$, where $x$-axis of $\zeta^2_a$ is the direction of motion of the agent at any time $t$. Vector $z_o$ is at $\pm{90}^0$ rotation with $z_p$. Let’s assume that at time $t$ there are certain number of obstacles surrounding the agent (and hence, ellipsoid $\Psi^t$). These obstacles are divided into two regions: those lying above $x$-axis and those lying below it in $\zeta^2_a$. If number of obstacles lying above is greater than that of below, then $z_o$ is at $-90^0$ w.r.t. $z_p$ and $+90^0$ (measured anticlockwise) otherwise. Let $z_p^u$ and $z_o^u$ be unit vectors in direction $z_p$ and $z_o$ respectively. A vector $z_e\in{R^2}, ||z_e||_2\leq{1}$ is then found by solving,
%
%
\begin{eqnarray}
\label{2d-ze}
\textbf{minimize}& &-(z_e)^Tz_p^u-\beta{log((z_e)^Tz_o^u)}\label{navigation_vector}\label{bicriterion}\\
\textbf{subject to}& &||z_e||_2\leq{1}\nonumber,
\end{eqnarray}
where $\beta\in(0,1]$ decides the trade-off in bi-criterion convex optimization problem. Fig-\ref{fig:inside_ellip_vectors} (left) depicts $z_p$, $z_o$ and $z_e$ for the 2D-problem.
%
%
\subsubsection{3D-ECAN: Avoiding Obstacles}
\label{3D-Navigation}

Since $P\in{S^3_{++}}$, its Eigenvectors are mutually orthogonal, forming a new coordinate system. Principal direction vector $z_p$ corresponds to the Eigenvector corresponding to $\lambda_{min}(P)$ and points towards $z^t_{g}$. The other two Eigenvectors are then rotated accordingly to form a right-hand coordinate system and they define the obstacle vectors $z^1_o$ and $z^2_o$. Assume that there are certain number of obstacles in the fov of the agent (and now they surround $\Psi^t$). Next, a vector (named collision-vector) is drawn from the center of ellipsoid to each of the obstacles and projection of these vectors is taken on $z^1_o$ and $z^2_o$. Fig-\ref{fig:inside_ellip_vectors} (right) depicts $z_p,z_o^1$ and $z^2_o$. Two parameters $s^+_1$ and $s^+_2$ are defined. If the number of collision-vector having negative projection on $z^1_o$ is greater than the number of collision-vector having positive projection on $z^1_o$, then $s^+_1=+1$, and $s^+_1=-1$ otherwise. In the same manner $s^+_2$ is computed, by taking projections on $z^2_o$. These two parameters, $s^+_1$ and $s^+_2$, effectively measure the number of obstacles in positive and negative $z^1_o$ and $z^2_o$ direction respectively. Let $z_{pu}, z^1_{ou}$ and $z^2_{ou}$ represent unit vectors in the direction $z_p,z^1_o$ and $z^2_o$ respectively. Also, let $\lambda_1$ and $\lambda_2$ are the Eigenvalues of $P$ corresponding to Eigenvectors to which $z^1_o$ and $z^2_o$ correspond, respectively. The resulting convex program for finding a vector $z_e\in{R^3},||z_e||_2\leq{1}$ can be compactly represented as,
%
%
\begin{eqnarray}
\label{3d-ze}
\textbf{minimize}& &-log\left(\frac{{(z_e^Tz^1_{ou})}^\frac{1}{\lambda_1}\times(z_e^Tz^2_{ou})^\frac{1}{\lambda_2}}{\exp\left\{\frac{-z_e^Tz_{pu}}{\lambda_{min}(P)}\right\}}\right)\nonumber\\
\textbf{subject to}& & ||z_e||_2\leq1.
\end{eqnarray}
Here trade-off is decided by inverse eigenvalues (axes-length of ellipsoid) which resolves the directional movement. Directions playing lesser important role for navigation to the goal location are given less importance. This not only helps in efficient navigation, but also avoids unnecessary directions, which is important in 3D-planning [7].
%
%
\subsubsection{Navigation Direction Inside 2D/3D-Ellipsoid}
\label{finiding-direction-zn}
Once $z_e$ is computed, a navigation direction $z_n$ for moving inside the ellipsoid can be found, and the procedure is, roughly speaking, the same for 2D and 3D space. A positive scalar $l_e$ is computed such that $z_b=(z_c+z_e\times{l_e})$ is a point on the  boundary of $\Psi^t$ with $z_c$ being its center. The navigation direction $z_n$ is a vector from $z^t_a$ to $z_b$, i.e. a vector from the current location of the agent inside the ellipsoid to the point $z_b$. Scalar $l_e$ can be computed as,
%
%
\begin{eqnarray}
\label{finding-le}
l_e&=&(-\Delta+\sqrt{\Delta^2-4\Lambda\Sigma})/({2\Lambda})\\
\Delta&=&2{z_c}^TP^t{z_e}+{z_e}^Tq^t;\text{ } \Lambda \text{ }=\text{ } {z_e}^TP^tz_e\nonumber\\
\Sigma&=&{z_c}^TP^t{z_c}+{z_c}^Tq^t+r^t\nonumber.
\end{eqnarray}
%
%
Once $z_n$ is computed, agent takes a small step $l_n$ in this direction, such that it still remains inside the ellipsoid. When the agent is point size, $l_n=\min(\delta_1,\delta_2)$, where $\delta_1$ is a predefined scalar, which may depend on speed of the agent or the step after which a re-computation of direction is required, or on the desired frequency of re-planning and $\delta_2=||z_a^t-z_b||_2$. However, when the agent has finite area/volume, then the length $l_n$ is computed by solving a quadratic program (or equivalently a SOCP). Fig-\ref{fig:table_f2} shows an instance when the agent has finite area, and the length $\delta_2$ is computed by solving the quadratic program (QP). Center of the agent is shown as red-dot in fig-\ref{fig:importance_f2}. The figure was meant to show the importance of function $f_2$. Next section explains complete framework for the finite-area/volume agent and also the QP for computing $l_n$. Fig-\ref{fig:inside_ellip_vectors} left and right show vectors $z_e$ (multiplied by $l_e$), and $z_n$ for 2D and 3D space respectively.
%
%
\section{ECAN: Finite-Size Agent \& Obstacle Architecture}
\label{finite-agent-architecture}

\begin{figure}[t]
\begin{center}
\subfloat[]{\label{fig:2d_agent}\includegraphics[trim =85mm 100mm 78mm 116mm, clip, width=1in]{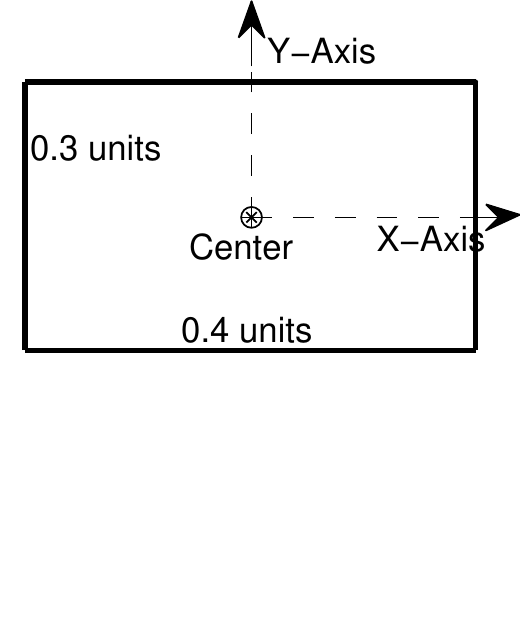}}
\subfloat[]{\label{fig:3d_agent}\includegraphics[trim = 47mm 92mm 47mm 92mm, clip, width=2.2in]{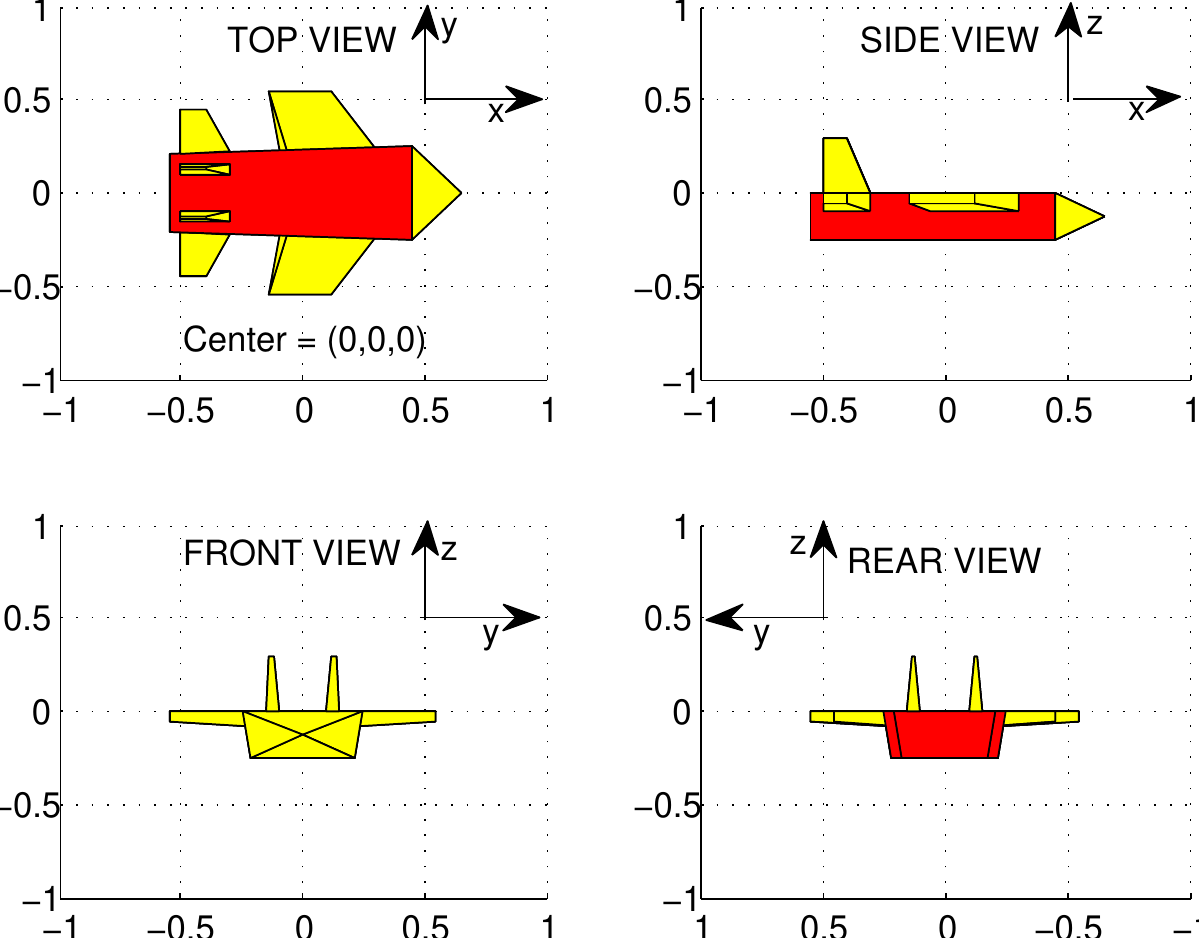}}
\end{center}
\caption{Finite area/volume agents used in the experiments: (a) shows a 2D-Agent with $x$-axis as the direction of motion; (b) shows a 3D-Agent with length of various parts described in the figure with the help of grid markings. }
\label{fig:agents}
\end{figure}

This section extends the idea behind the point-agent model to the finite-area/volume agent avoiding the finite-area/volume obstacles. Section-\ref{obstacle-sampling} explains the point-cloud representation of the finite-obstacles lying in fov of the agent. The finite agents, for 2D and 3D spaces, used in the experiments are shown in fig-\ref{fig:2d_agent} and fig-\ref{fig:3d_agent} respectively. Convex constraints in (\ref{ccf1f2f3}) kept the point-agent inside the resulting ellipsoid, same concept applies for the finite agent. Utilizing the property that a convex set $\mathbf{C}_1$ contains a set $\mathbf{C}_2$ (need not be convex), iff $\mathbf{C}_1$ contains the convex-hull of $\mathbf{C}_2$. Therefore the finite agents (where the 2D-agent forms a convex set while the 3D-agent forms a non-convex set) can be constrained to lie inside the ellipsoid by constraining the $m$-extremum locations on these agents to lie inside the ellipsoid $\Psi^t$. The extremum points for the 2D agents are the 4 corner points ($m=4$), while for the 3D-agent (a plane), the extremum points are the corner points of the free end of front, rear and top wings (all are yellow in fig-\ref{fig:3d_agent}); 4 corner points of the front and back ends of the central body (red in fig-\ref{fig:3d_agent}); and the tip of the nose of the plane (font yellow colored tetrahedron in fig-\ref{fig:3d_agent}), leading to a total of $m=33$ locations on the plane. Thus the first constraint in convex-program (\ref{ccf1f2f3}) is replaced with the constraint ${z^t_{a}(i)}^TP^tz^t_a(i)+{z^t_a(i)}^Tq^t+r^t\leq{-1}, i={1,...,m}$; $z^t_a(i)$ denotes $i^\text{th}$ location on the agent's body. Also, when the agent is finite-size, variable $z^t_a$ represents its center at time $t$. Therefore, function $f_2$ in (\ref{ccf1f2f3}) now minimizes the distance of center of the agent with the boundary of resulting ellipsoid. The basic working principle of ECAN planning remains the same and only the step length $l_n$ for moving inside the ellipsoid, in the direction $z_n$, changes. Step-length $l_n=\min(\delta_1,\delta_2)$, where $\delta_1$ is a predefined scalar quantity, and $\delta_2$ is found by solving a QP (next section) by constraining the agent to remain inside the ellipsoid $\Psi^t$.
%
%
%
%
\subsection{Convex Quadratic Programming for Finding $l_n$}
\label{Quadratic-Program}
Let there be a data-structure $D^t_a$ that stores location of different points on the agent's body in the form of vectors in agent's local axis system ($x$, $y$ and $z$ directions), as shown in fig-\ref{fig:agents} for both 2D and 3D agents, w.r.t to center of the agent ($z^t_a$). It also stores agent's local axis system according to the agent's orientation in space at time $t$. Therefore whatever be the orientation of the agent in space is, location of the points on the agent's body is given by $z^t_a(i) = z^t_a + D^t_a(i)$, where $D^t_a(i)$ is the vector (depends on the current orientation of the agent) for the $i^\text{th}$ location on the agents body w.r.t the agent's center $z^t_a$. Therefore the variable $\delta_2$ (and hence $l_n$), at time $t$, for moving inside the ellipsoid, in direction $z_n$, can be found by solving following Convex-QP (with variable $x$ and $\delta_2$):
%
%
\begin{eqnarray}
\label{finding-delta2}
&&\textbf{maximize}\text{  }\text{ } \delta_2\\
&&\textbf{subject to}\text{ }\text{ }x=\delta_2{z_n}; \text{ } \text{ }\text{ }i=\{1,...,m\}\nonumber\\
&&\text{  }x^TP^tx\leq{-(2P^tD^t_a(i)'}+q^t)^Tx+\Psi^t(D^t_a(i)')-1\nonumber.
\end{eqnarray}
%
%
The constraints are obtained by expanding the ellipsoid constraint $\Psi^t(z^t_a+z_n\times\delta_2+D^t_a(i))<=-1$, for each of the $m$ extremum locations and using $D^t_a(i)'=z_a^t+D^t_a(i)$. After solving the above QP, step-length can be found as $l_n=\min(\delta_1,\delta_2)$. Complete ECAN algorithm is explained as Algorithm-(\ref{ECAN-Algorithm}). Input parameters for subroutines are defined inside the curly braces, with optional input parameters further inside the square braces. Sub-routine \textit{getObstaclesInFov} identifies obstacles in fov of the agent at time $t$; \textit{getEllipsoid} forms the ellipsoid by solving-(\ref{ccf1f2f3}); \textit{getDirectionInEllipsoid} finds direction $z_e$ using optional input $\beta$ (active) for 2D navigation by solving-(\ref{2d-ze}) and by solving convex program-(\ref{3d-ze}), without $\beta$, for 3D-navigation; \textit{solveAnalyticEquation} finds $l_e$ by using-(\ref{finding-le}); \textit{getMotionDirection} finds $z_n$ as mentioned in section-\ref{finiding-direction-zn}; \textit{isPointAgent} is a Boolean variable which is \textit{true} for the point-agent and \textit{false} otherwise; and \textit{getLengthSolveQP} finds $\delta_2$ by solving-(\ref{finding-delta2}).
%
%
%
\begin{algorithm}[t]                      
\caption{-Ellipsoidal Constrained Agent Navigation}          
\label{ECAN-Algorithm}                           
\begin{algorithmic}                    
\STATE - initialize parameters: $\delta_1,\gamma,\alpha,\beta$ and $\epsilon$ (small constact)
\STATE - initialize variables: $V_{fov}=(R_\text{fov},\theta_\text{fov},\phi_\text{fov},dr,d\theta,d\phi)$
\STATE - initialize agent (at $t=0$) and goal location: $z^0_a, D^0_a, z_g$
\STATE {\,\,} \textbf{while} ($||z^t_a-z_g||_2>\epsilon$)
\STATE {} \text{ } \text{ } $O^t\leftarrow$ \textit{getObstaclesInFov}$\{z^t_a,D^t_a,V_\textit{fov}\}$
\STATE {} \text{ } \text{ } $\Psi^t\leftarrow$ \textit{getEllipsoid}$\{z^t_a,O^t,z_g,D^t_a,\gamma,\alpha\}$
%
%
\STATE {} \text{ } \text{ } \textbf{if} ($|\Psi^t(z_g)|>\epsilon$)
\STATE {} \text{ } \text{ } \text{ } \text{ } \text{ } $z_e\leftarrow$ \textit{getDirectionInEllipsoid}$\{\Psi^t,D^t_a,O^t,[\beta]\}$
\STATE {} \text{ } \text{ } \text{ } \text{ } \text{ } $l_e\leftarrow$ \textit{solveAnalyticEquation}$\{\Psi^t,z_e\}$
\STATE {} \text{ } \text{ } \text{ } \text{ } \text{ } $z_n\leftarrow$ \textit{getMotionDirection}$\{z_e,l_e,\Psi^t\}$ 
\STATE {} \text{ } \text{ } \text{ } \text{ } \text{ } \textbf{if} (\textit{isPointAgent})
\STATE {} \text{ } \text{ } \text{ } \text{ } \text{ } \text{ } \text{ } \text{ } $l_n\leftarrow\min(\delta_1,||z^t_a-z_g||_2)$
\STATE {} \text{ } \text{ } \text{ } \text{ } \text{ } \textbf{else}
\STATE {} \text{ } \text{ } \text{ } \text{ } \text{ } \text{ } \text{ } \text{ } $\delta_2\leftarrow$ \textit{getLengthSolveQP}$\{z^t_a,z_n,\Psi^t,D^t_a\}$
\STATE {} \text{ } \text{ } \text{ } \text{ } \text{ } \text{ } \text{ } \text{ } $l_n\leftarrow\min(\delta_1,\delta_2)$
\STATE {} \text{ } \text{ } \text{ } \text{ } \text{ } \textbf{endif}
\STATE {} \text{ } \text{ } \textbf{else}
\STATE {} \text{ } \text{ } \text{ } \text{ } \text{ } $z_n\leftarrow(z_g-z^t_a)/(||z_g-z^t_a||_2)$ 
\STATE {} \text{ } \text{ } \text{ } \text{ } \text{ } \textbf{if} (\textit{isPointAgent})
\STATE {} \text{ } \text{ } \text{ } \text{ } \text{ } \text{ } \text{ } \text{ } $l_n\leftarrow\min(\delta_1,||z^t_a-z_g||_2)$
\STATE {} \text{ } \text{ } \text{ } \text{ } \text{ } \textbf{else}
\STATE {} \text{ } \text{ } \text{ } \text{ } \text{ } \text{ } \text{ } \text{ } $\delta_2\leftarrow$ \textit{getLengthSolveQP}$\{z^t_a,z_n,\Psi^t,D^t_a\}$
\STATE {} \text{ } \text{ } \text{ } \text{ } \text{ } \text{ } \text{ } \text{ } $l_n\leftarrow\min(\delta_1,\delta_2)$
\STATE {} \text{ } \text{ } \text{ } \text{ } \text{ } \textbf{endif}
\STATE {} \text{ } \text{ } \textbf{endif}
\STATE {} \text{ } \text{ } $z^{t+1}_a\leftarrow z^t_a + z_n\times{l_n}$
\STATE {} \text{ } \text{ } $t\leftarrow{t+1}$
\STATE  \textbf{endwhile}
\end{algorithmic}
\end{algorithm}
%
%
\subsection{Finite Area/Volume Obstacles}
\label{obstacle-sampling}

This section now incorporates finite obstacles for developing an architecture that handles finite-agent and finite-obstacles using convex formulation described in earlier sections. As described in sec-\ref{basis-underlying-priniciple}, agent's fov is restricted by $(R_\textit{fov},\theta_\textit{fov},\phi_\textit{fov})$. To incorporate finite obstacles, fov is discretized. The discretization parameters $(dr,d\theta,d\phi)$ represent discretization of respective components. The number of discretized grid-points is $N=(2\theta_\textit{fov}+1)(2\phi_\textit{fov}+1)R_\textit{fov}/(dr.d\theta.d\phi)$ for 3D space and $N=(2\theta_\textit{fov}+1)R_\textit{fov}/(dr.d\theta)$ for 2D space. This discretization is done in the agent's local coordinate system, and therefore needs to be done only once throughout the navigation. At any time $t$, these points are mapped to global coordinate system to get global coordinates of these discretized grid-points. At time step $t$, once the agent finds any finite-obstacle in its fov, the grid-points lying on the body of these obstacles are marked as point-obstacles in the environment. The \textit{getObstaclesInFov} subroutine uses these discretization parameters to convert the part of the finite-obstacles lying in the agent's fov to a point-cloud. Also, this discretization is done only for simulation purposes. In the real-world implementation these points can be extracted from the output of any perception sensor, for example, coordinates of pixels defining the obstacles in case of image based perception, or using LADAR sensors ([8],[14]).
%
%
%
%
\section{Experiments}
\label{experiments}
This section lists various experiments to demonstrate performance of ECAN in cluttered environments, both using the point and finite obstacle. The purpose of including point obstacle is to increase the random clutterness in the environment. Finite agents used in the experiments for 2D and 3D navigation are shown in fig-\ref{fig:agents}. Default values of parameters are: $(\alpha,\beta,\epsilon,R_{fov},\phi_\textit{fov},d\phi)=(0.1,1,10^{-2},5,40^0,0.5^0)$; $\gamma=5\times{10^{-5}}$ when only point obstacles are present, and $\gamma=5\times{10^{-4}}$ otherwise; and $(\delta_1,\theta_\textit{fov},dr,d\theta)=(1,80^0,0.2,1^0)$ \& $(2,40^0,0.1,0.5^0)$ for 2D \& 3D navigation respectively. In all the experiments, environment is completely unknown and unseen by the agent, and also the agent has no knowledge about the environment beyond its field-of-view.
%
%
\begin{figure}[t]
\begin{center}
\includegraphics[trim =53mm 96mm 48mm 92mm, clip, width=7cm]{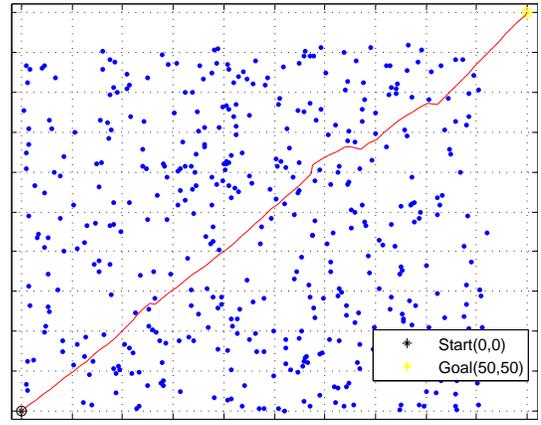}
\end{center}
\caption{2D-navigation: point obstacles with finite agent.}
\label{fig:exper2dpt}
\end{figure}
%
\begin{figure}[t]
\begin{center}
\includegraphics[trim =53mm 96mm 48mm 92mm, clip, width=7cm]{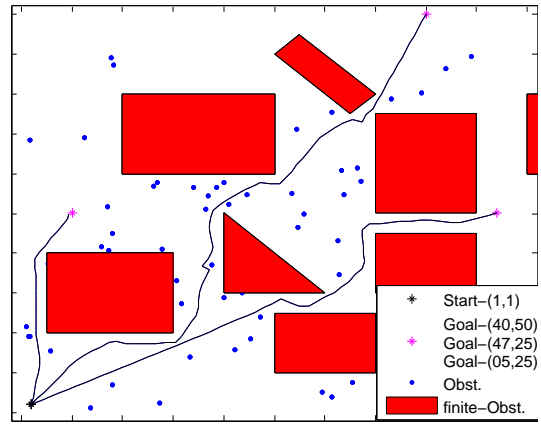}
\end{center}
\caption{Three experiments for 2D-Navigation among finite and random-point obstacles with finite-agent.}
\label{fig:2dmixed}
\end{figure}
%
%
\subsection{2D-Navigation: Point Obstacels with Finite Agent}
This experiment demonstrates ECAN's ability to plan a path and discover gaps, through which the finite-agent can pass, in randomly cluttered environment with point-mass obstacles, to reach the goal-location. Fig-\ref{fig:exper2dpt} shows a resulting path with randomly generated 446-point obstacles. At $t=0$ agent is facing towards the goal location.%
%
%
\subsection{2D-Navigation: Finite-Agent \& Mixed Obstacles}
Three experiments shown in fig-\ref{fig:2dmixed} demonstrate navigation with ECAN to 3-different locations with same initial conditions. These experiments demonstrate that the point obstacle representation of the boundary of finite obstacles is sufficient for ECAN to avoid any kind of finite obstacles. A spline can be fit ([21]), to further get a smooth path inside the ellipsoid (to avoid sudden turns), by considering the agent's currnet location inside the ellipsoid, ellipsoid's center and the point $z_b$ on the boundary of ellipsoid.
%
%
\begin{figure}[t]
\begin{center}
\subfloat[]{\label{fig:3d-Nav-3d}\includegraphics[trim =53mm 99mm 48mm 93mm, clip, width=7cm]{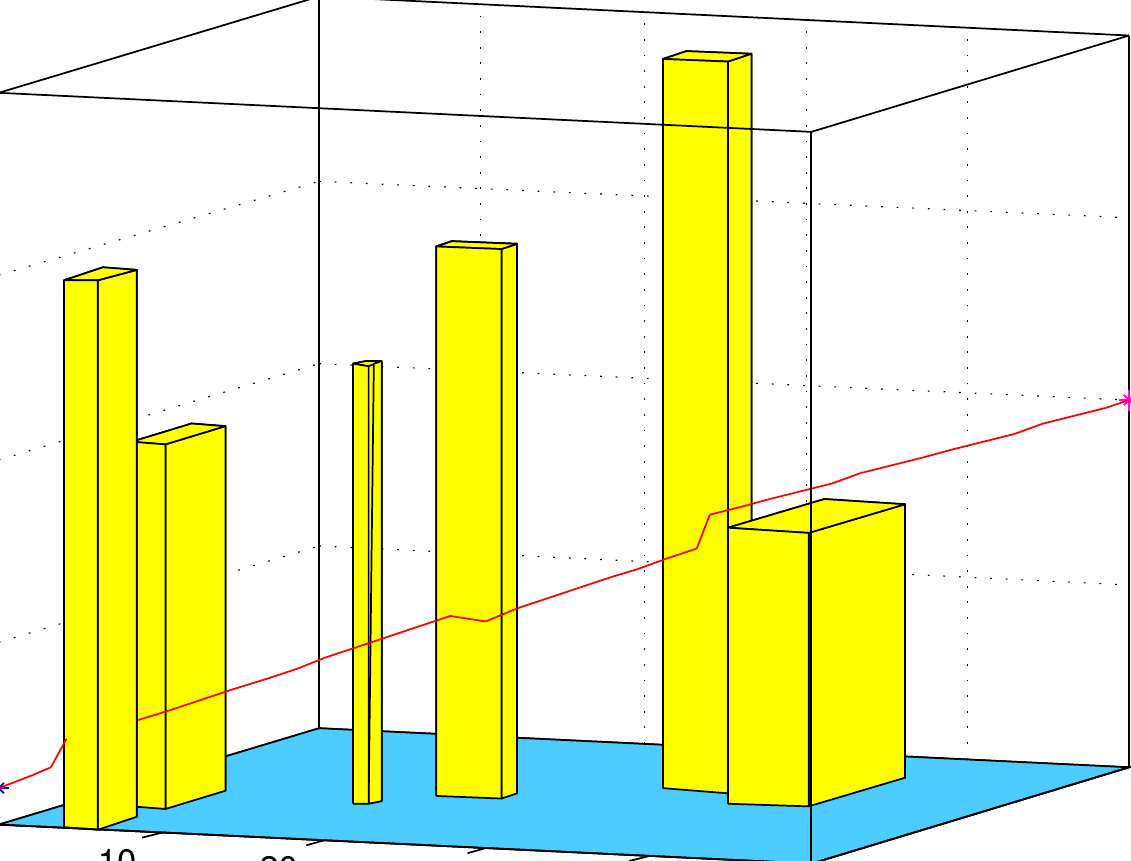}}\\
\subfloat[]{\label{fig:3d-NavProj}\includegraphics[trim =53mm 96mm 47mm 92mm, clip, width=4.5cm]{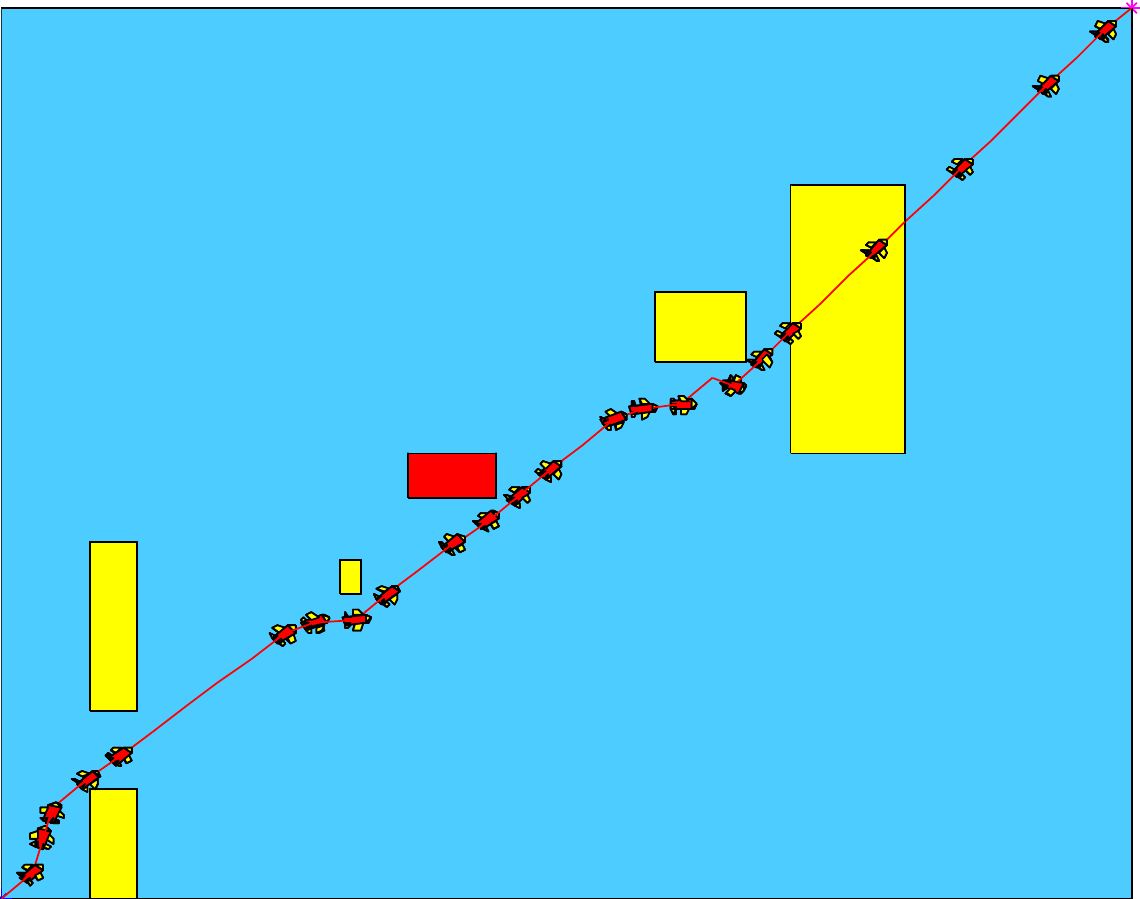}}
\subfloat[]{\label{fig:3d-NavMag}\includegraphics[trim =34mm 50mm 50mm 90mm, clip, width=2.5cm]{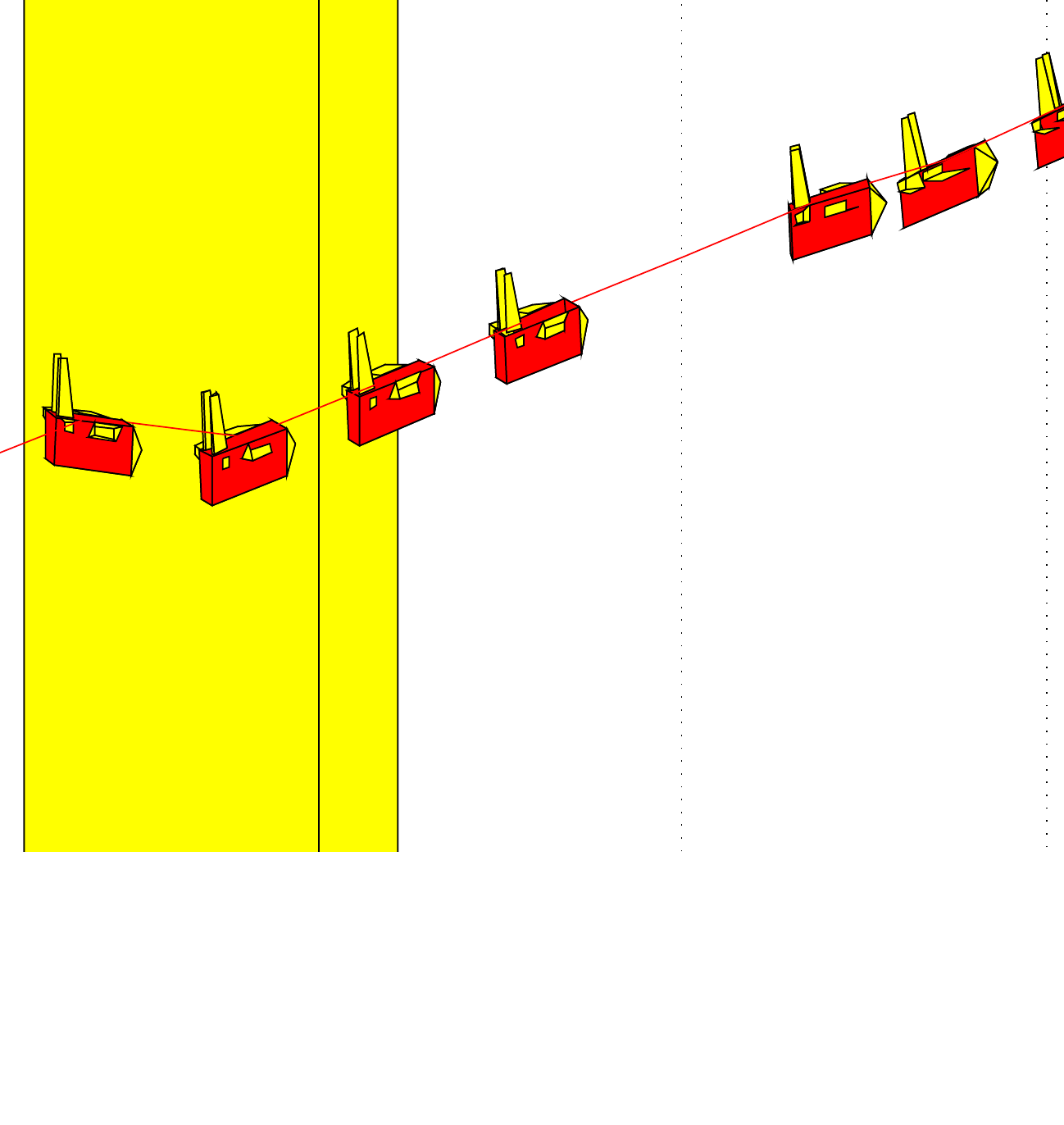}}
\end{center}
\caption{3D-Navigation with finite agent; $z^0_a=(0,0,1)$ and $z_g=(50,50,10)$.}
\end{figure}
%
%
\subsection{3D-Navigation: Finite-Agent \& Finite Obstacles}
The 3D-agent (a fighter plane shaped UAV) navigates successfully in completely unknown and unseen environment using the finite-agent architecture of 3D-ECAN algorithm. To avoid clutterness, and to show clearly the path planned in online fashion, agent is not shown in fig-\ref{fig:3d-Nav-3d}. Fig-\ref{fig:3d-NavProj} depicts projection on $(x,y)$-plane and non-vertical turns to avoid obstacles. It shows the agent at key locations, such as when it is elevating and making turns. A magnified image of the agent is shown in fig-\ref{fig:3d-NavMag}, while avoiding the red-top building (fig-\ref{fig:3d-NavProj}). At $t=0$ agent is facing at $(1,1,0.5)$.
%
%
\begin{figure}[t]
\begin{center}
\includegraphics[trim =42mm 87.5mm 39.5mm 92mm, clip, width=7cm]{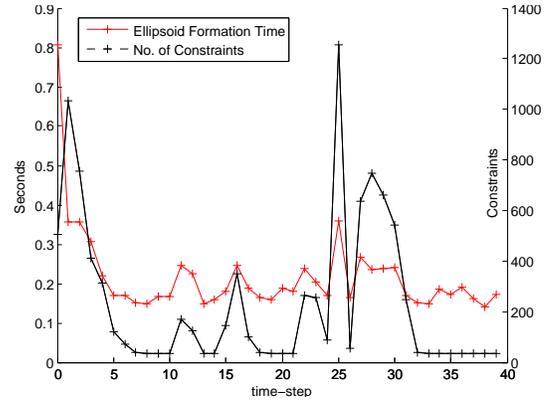}
\end{center}
\caption{3D-Navigation: Ellipsoid formation time and number of constraints at each time step $t$; $t=\{0,...,39\}.$}
\label{fig:timeellipsoid}
\end{figure}
%
%
\subsection{Experimental Analysis of Computational Time}
This section experimentally analyzes computation time required by the convex programs. ECAN was implemented on MATLAB-R2008a using non-commercial CVX ([22]) on Sony Vaio Laptop, $2.66$-GHz Core i5-480M, $4$-GB RAM, running Windows-$7$ $64$-bit. First, the case of 2D-finite obstacle with finite-agent is examined. These results are shown specifically for the path leading to (40,50) in fig-\ref{fig:2dmixed} and conforms with the average result over several random 2D-domains. The total planning steps is 72, i.e. $t=\{0,...,71\}$. The average time taken by convex program-(\ref{ccf1f2f3}) is $0.15963$-second and standard deviation $0.01997$ with number of constraints varying from $6$ to $130$ including the positive-definite constraint on $P^t$; average time taken by convex problem-(\ref{finding-delta2}) is $0.36788$-second over $39$ calls to the program; and the average time by problem-(\ref{finding-le}) is $0.0933$-second over 72 calls. For 3D-path planning shown in fig-\ref{fig:3d-Nav-3d}, time taken to form the ellipsoid at each time-step along with the number of constraints in the QCQP is shown in fig-\ref{fig:timeellipsoid}. The excessive time taken at $t=0$ is due to the close proximity of the agent to ground (at $t=0$ and hence point obstacles) and with a wall standing in front, with length and width of the agent taken into account, it restricts the ellipsoid by a great amount. For all other steps, it can be seen that the ellipsoid adjusts itself accordingly, thereby reducing the computation time. The average time taken for convex problem-(\ref{finding-delta2}), over 39 calls, to find $\delta_2$ is $0.09097$-second, with maximum time of $0.12082$-second at $t=0$; average time taken by convex program-(\ref{finding-le}) to find $z_e$ is $0.4068$-second and was called $12$ times.

\section{Discussion, Conclusion \&  Future Work.}
This paper presented a Convex-QCQP based $3$-stage convex-optimization algorithm ECAN for online path planning in completely unknown and unseen continuous environments. The algorithm has apparent computational advantage over integer-programming based path planning (see [11],[12] for comparison), and moreover is able to plan in unseen spaces, which most of the integer-program based path planning algorithms can't. However, a key point that is not addressed in this paper for 3D-planning is to determine the rotation, about local $x$-axis, of the 3D-agent according to the eigenvector-coordinate system of the ellipsoid. The 3D-ellipsoid not only provides a navigation direction, but also there's an important information about how the agent should rotate itself to nagivate as far as possible inside the ellipsoid. This can be particuarly useful for fixed-wing UAV navigation (high-speed), as it provides UAV with the path and also with information about orientation to avoid obstacles. Also, dynamics of the vechile and minimum turn radius constraints were not addressed while planning the path. While determining $z_e$, an extra constraint of maximum deviation from current motion direction can be added that takes into account the turn radius. Addressing these areas requires a separate full length paper and will be addressed in future.

We would also like to comment over \textit{unseen spaces}, emphasized thoughout the paper. The term \textit{unknown spaces} has been used widely in the literature, specifically in the research using artificial intelligence and some of the references mentioned in the related work. Most literature first propose a method to plan in unknwon spaces, but eventually they build a method that first either captures model of the environment or requires a path from the agent's initial location to the goal location, and then path planning occurs. This makes the environment (partially or completely) known. Therefore, we emphasized \textit{unseen spaces} to clearly specify that the agent is unaware of upcoming geometrical-hindrances and obstacles as it navigates through the completely unknwon environment.

A key advantage and the reson behind success of ECAN in \textit{unseen spaces} is that the orientation of ellipsoid intelligently incorporates the direction leading to a goal location. This provides robustness to the ECAN in finding a feasible path (if it exists) to the goal location, no matter how cluttered the space is. Also, convex formulation guarantees an optimal solution at each time-step and makes ECAN computationally tractable.


[1] S. Boyd and L. Vandenberghe, {\it Convex Optimization} (Cambridge University Press, 2004).\smallskip\newline
[2] S. Sharma, E.A. Kulczycki and A. Elfes, Trajectory Generation and Path Planning for Autonomous Aerobots, {\it ICRA Workshop on Robotics in Challenging and Hazardous Environments}, 2007.\smallskip\newline
[3] K. Yang and S. Sukkarieh, Planning Continuous Curvature Paths for UAVs Amongst Obstacles, {\it In Australasian Conference on Robotics and Automation}, 2008.\smallskip\newline
[4] A. Stentz, The focussed D* algorithm for real-time replanning, {\it In Int. Joint Conference on Artificial Intelligence}, pp. 1652–1659, 1995. \smallskip\newline
[5] S. Koenig and M. Likhachev, Fast Replanning for Navigation in Unknown Terrain, {\it IEEE Transactions on Robotics}, 21(3), 2005.\smallskip\newline
[6] D. Ferguson and A. Stentz, Field D*: An Interpolation-based Path Planner and Replanner, {\it In Int. Sysposium on Robotics Research}, 2005.\smallskip\newline
[7] J. Carsten, D. Ferguson and A. Stenz, 3D Field D: Impoved Path Planning and Replanning in Three Dimensions. {\it IEEE/RSJ Int. Conference on Intelligent Robots and Systems}, 2006.\smallskip\newline
[8] D. Dolgov, S. Thrun, M. Montemerlo and J. Diebel, Path Planning for Autonomous Vehicles in Unknown Semi-structured Environments, {\it Int. Journal of Robotics Research}, 2010.\smallskip\newline
[9] A. Richards, Y. Kuwata, and J. How, Experimental Demonstrations of Real-time MILP Control, {\it AIAA Guidance, Navigation, and Control Conference}, 2003.\smallskip\newline
[10] T. Schouwenaars, J. How and E. Feron, Receding Horizon Path Planning with Implicit Safety Guarantees, {\it In American Control Conference}, Boston, M.A., 2004.\smallskip\newline
[11] T.A. Ademoye, A. Davari, C.C. Caltello, S. Fan, and J. Fan, Path Planning via CPLEX Optimization, {\it Southeastern Symposium on Systems Theory}, University of New Orleans, March 16-18, 2008.\smallskip\newline
[12] M.P. Vitus, V. Pradeep, G.M. Hoffmann, S.L. Waslander and C.J. Tomlin, Tunnel-MILP: Path Planning with Sequential Convex Polytopes, {\it Proc. AIAA Guidance, Navigation and Control Conference}, 2008.\smallskip\newline
[13] Lars Blackmore, Robust Path Planning and Feedback Design under Stochastic Uncertainty, In {\it Proc. of AIAA Guidance, Navigation and Control Conference}, 2008.\smallskip\newline
[14] Nicolas Vandapel, James Kuffner, Omead Amidi, Planning 3-D Path Networks in Unstructured Environments, {\it Proc. Int. Conference on Robotics and Automation}, 2005.\smallskip\newline
[15] J.J. Kuffner, Efficient optimal search of euclidean-cost grids and lattices, {\it In IEEE/RSJ Int. Conference on Intelligent Robots and Systems}, 2004.\smallskip\newline
[16] J. Miura, Support Vector Path Planning, {\it In IEEE/RSJ Int. Conference on Intelligent Robots and Systems}, 2006.\smallskip\newline
[17] O. Hachour, A three-dimensional collision- free-path planning, {\it Int. Journal of Systems Applications, Engineering \& Development}, 4(3), 2009.\smallskip\newline
[18] T.-K. Wang, Q. Dang and P.-Y. Pan, Path Planning Approach in Unknown Environment, {\it Int. Journal of Automation and Computing}, 4(3), 310-316, 2010.\smallskip\newline
[19] T. Schouwenaars, M. Valenti, E. Teron and J. How, Implementation and Flight Test Results of MILP-based UAV Guidance, {\it Proc. of the IEEE Aerospace Conference}, 2005.\smallskip\newline
[20] Y. Kuwata and J. How, Three Dimensional Receding Horizon Control for UAVs, {\it AIAA Guidance Navigation and Contro Conference and Exhibit}, 2004.\smallskip\newline
[21] T. Ersson and X. Hu, Path Planning and Navigation of Mobile Robots in Unknown Environments, {\it In IEEE/RSJ Int. Conference on Intelligent Robots and Systems}, 2001.\smallskip\newline
[22] CVX: MATLAB software for disciplined convex porgramming by M. Grant and S. Boyd: {http://cvxr.com/cvx/}
\end{document}